\documentclass[runningheads]{llncs}
\usepackage[T1]{fontenc}
\usepackage[table]{xcolor}
\usepackage{amsmath}
\usepackage{amssymb}
\usepackage{graphicx}
\usepackage{tabularx}
\usepackage{todonotes}
\usepackage{booktabs}
\usepackage{layouts}

\newcommand{\reffig}[1]{Fig.~\ref{#1}} 
\newcommand{\reftab}[1]{Tab.~\ref{#1}} 
\newcommand{\refeq}[1]{Eq.~\ref{#1}}  

\newcommand{\xcont}{\mathbf{X}_{cont}}
\newcommand{\xint}{\mathbf{X}_{int}}
\newcommand{\xcat}{\mathbf{X}_{cat}}

\definecolor{ForestGreen}{RGB}{34,139,34}

\usepackage{orcidlink}
\renewcommand{\orcidID}[1]{\raisebox{1.1ex}{\orcidlink{#1}}}

\begin{document}
\title{Hybridizing Target- and SHAP-encoded Features for Algorithm Selection in Mixed-variable Black-box Optimization}
\titlerunning{Hybridizing TE and SH Features for AS in Mixed-variable BBO}  
%
\author{Konstantin Dietrich\inst{1,2}\orcidID{0000-1111-2222-3333} \and
Raphael Patrick Prager\inst{3}\orcidID{0000-0003-1237-4248} \and
Carola Doerr\inst{4}\orcidID{0000-0002-4981-3227}
\and Heike Trautmann\inst{5,6}\orcidID{0000-0002-9788-8282}}
\authorrunning{K. Dietrich et al.}

\institute{Big Data Analytics in Transportation, TU Dresden, Germany \and
ScaDS.AI, Dresden, Germany\\
\email{konstantin.dietrich@tu-dresden.de}
\and
Data Science: Statistics and Optimization, University of Münster, Germany\\
\email{raphael.prager@wi.uni-muenster.de}
\and
Sorbonne Universit\'e, CNRS, LIP6, Paris, France\\
\email{carola.doerr@lip6.fr}
\and
Machine Learning and Optimisation, Paderborn University, Germany
\email{heike.trautmann@uni-paderborn.de}
\and
Data Management and Biometrics Group, University of Twente, Netherlands}

\maketitle              
\begin{abstract}
Exploratory landscape analysis (ELA) is a well-established tool to characterize optimization problems via numerical features. ELA is used for problem comprehension, algorithm design, and applications such as automated algorithm selection and configuration. Until recently, however, ELA was limited to search spaces with either continuous or discrete variables, neglecting problems with mixed variable types. This gap was addressed in a recent study that uses an approach based on target-encoding to compute exploratory landscape features for mixed-variable problems. 

In this work, we investigate an alternative encoding scheme based on SHAP values. While these features do not lead to better results in the algorithm selection setting considered in previous work, the two different encoding mechanisms exhibit complementary performance. Combining both feature sets into a hybrid approach outperforms each encoding mechanism individually. Finally, we experiment with two different ways of meta-selecting between the two feature sets. Both approaches are capable of taking advantage of the performance complementarity of the models trained on target-encoded and SHAP-encoded feature sets, respectively.

\keywords{Mixed-Variable Optimisation  \and SHAP \and Automated Algorithm Selection}
\end{abstract}

\section{Introduction}

Exploratory landscape analysis (ELA), also referred to as fitness landscape analysis, has shown to be a powerful tool to characterize the landscape of continuous black-box optimization problems~\cite{Mersmann2011}. This type of problems lacks a closed-form representation and does not provide any gradient information to exploit. 

In a plethora of studies, ELA's usefulness has been demonstrated; either to better understand the behavior of an algorithm during its optimization process~\cite{lunacek2006,trajectoryFeatures2022}, to determine the degree of multi-modality, global structure or other structural properties of an optimization landscape~\cite{Mersmann2011,SeilerPKT2022ACollection}, or to use ELA as a means in automated algorithm selection~\cite{AASsmithmiles2009,AASmunoz2015,KerschkeT2019AutomatedAlgorithm,KerschkeHNT2019AutomatedAlgorithm} and configuration~\cite{BelkhirDSS17,GuzowskiS23,PikalovM21}.
Despite this popularity, ELA has neither been disseminated to other research domains, nor is it really used by practitioners. 
We hypothesize that one major caveat is the limitation to the purely continuous space. In fact, many black-box optimization problems exhibit a mixed-search space, where the search space can be an arbitrary mixture of continuous, integer, and categorical decision variables. This type of problems is often referred to as \textit{mixed-variable problem} (MVP)~\cite{lucidi05mvp2,pelamatti18mvp}.
While there have been successful attempts to extend ELA to the binary or mixed-integer space~\cite{wmodel,prager23mip}, these works still do not consider categorical decision variables.

This has been addressed in a recent study by~\cite{prager2024mvp}. The authors propose a methodology which allows the computation of ELA features for black-box problems with mixed search spaces. This methodology is evaluated in an automated algorithm selection setting where several algorithms are benchmarked on a set of \textit{hyperparameter optimization} (HPO) problems as important representatives of the MVP domain. Thereafter, ELA features are used to automatically select an appropriate algorithm out of the portfolio for any given problem instance. Thereby, they demonstrate that their developed methodology is able to discriminate between problem instances. The results show that this approach is superior compared to relying on any single algorithm out of the portfolio.
Their proposed methodology contains a variety of different steps of which one pertains to the transformation of categorical variables into continuous ones. The considered transformations are \textit{one-hot encoding} (as a baseline) and \textit{target-encoding}~\cite{te2001}. 

The contribution of this study is that we extend the work of~\cite{prager2024mvp} by introducing a different encoding method based on SHAP (SHapley Additive exPlanations) values~\cite{NIPS2017_7062}. These values have the advantage that they can represent the features additive contribution to the prediction. In doing so the categorical features are encoded on a `per observation'-basis while taking into account the interactions to all other contributing features. In this scenario an observation is a single row of data containing various features (input variables) and their corresponding target values (output variable).
We use the same experimental settings as~\cite{prager2024mvp} to ensure a fair comparison and highlight the merits as well as the caveats of our transformation variant. Our devised method does not outperform the existing results of~\cite{prager2024mvp} across all considered problem instances, yet it performs well in certain areas. A combination or what we call hybridization of both encoding methods produces a superior performance than any encoding method achieves on its own. Leveraging two sophisticated approaches to combine both encoding methods, we manage to improve the automatic algorithm selection substantially.


\section{Mixed-Variable Black-Box Optimization}\label{sec:mvp}
Many optimization problems and HPO problems in particular have a mixed search space. Meaning, the search space of these problems is constituted by a mixture of continuous ($\xcont$), integer ($\xint$), and categorical decision variables ($\xcat$).
The latter type poses two challenges for the optimization community especially. First, categorical decision variables do not possess any inherent order relation. 
Furthermore, $\xint$ and $\xcat$ can impose hierarchical/conditional structure onto other continuous decision variables. In other words, a specific instance of a given categorical decision variable can govern the domain of $\xcont$. In~\cite{prager2024mvp}, the authors purposefully ignore these constraints in their preprocessing scheme for ELA feature computation. They make the case that the infeasible regions of the search space (reached by ignoring these constraints) are $n$-dimensional hyperplanes with a constant objective value in the direction of the decision variables with the violated constraint. This information is exploited in their feature generation process. It is important to note that the constraints are only relaxed in the ELA feature generation process and are still in place when algorithms are applied to solve the problem in question.

Hence, we adopt the simplified formal representation of an MVP of~\cite{prager2024mvp}:
\begin{align}\label{eq:mvp}
    \begin{split}
        \text{min}\quad & f(\xcont, \xint, \xcat) \\
        \text{w.r.t.}\quad & \xcont \in  \mathbb{R}^{n_{cont}} \\ 
        & \xint \in \mathbb{Z}^{n_{int}} \\
        & \xcat \in \mathbb{Z}^{n_{cat}} \\
        \text{s.t.}\quad & \mathbf{g}(\xcont, \xint, \xcat) = 0 \\
        & \mathbf{h}(\xcont, \xint, \xcat) \leq 0,
    \end{split}
\end{align}
where $f$ denotes the objective function with its respective equality constraints $\mathbf{g}$ and inequality constraints $\mathbf{h}$.
With that, we define the decision space of our MVP as $\mathbf{X} = (\xcont, \xint, \xcat)$ and the objective space is denominated as $\mathbf{Y}$.

\section{Problem Representation}\label{sec:prob-repr}
\subsection{Exploratory Landscape Analysis}\label{sec:ela}

\begin{figure}
    \centering
    \includegraphics[width=\textwidth]{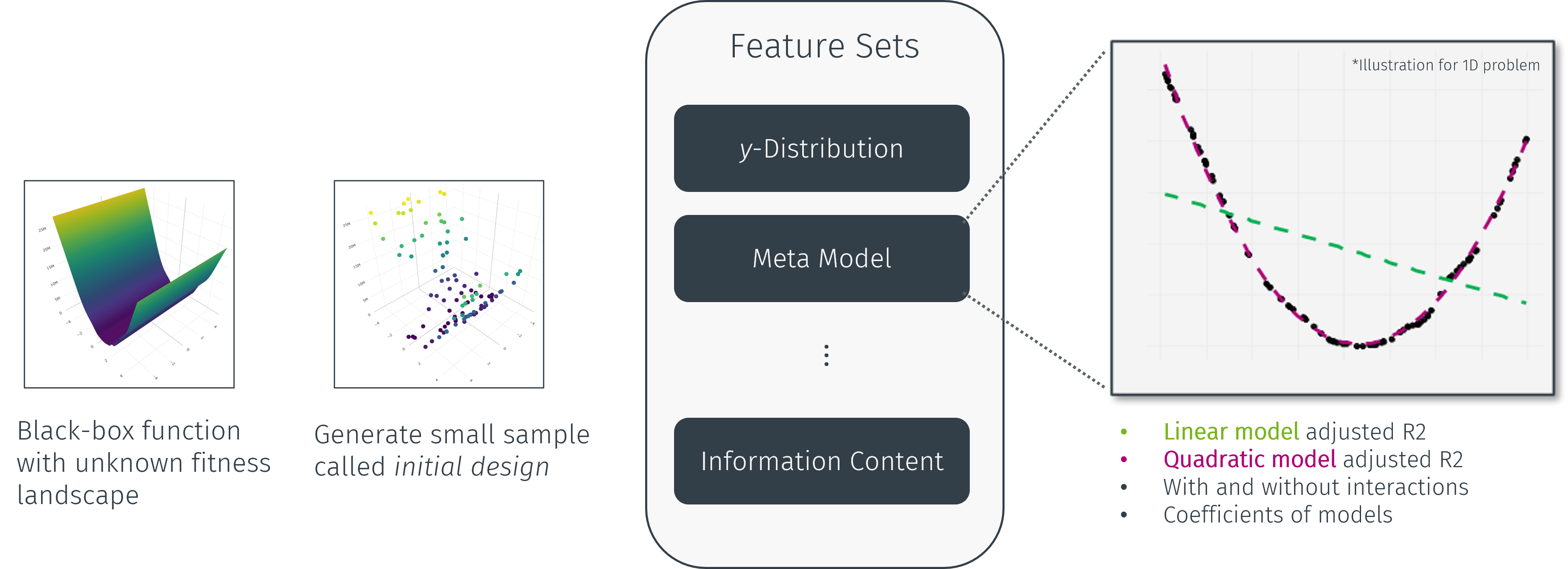}
    \caption{General procedure of ELA feature generation. Based on the initial design several feature sets can be computed. The feature set `meta model' fits the several linear and quadratic models and uses model coefficients as well as the adjusted $R^2$ as features.}
    \label{fig:ela}
\end{figure}
Black-box problems in general do not offer any insight into their landscape and what structures they are composed of. To provide the means to distinguish between and characterize different continuous single-objective black-box optimization problems, \textit{exploratory landscape analysis} (ELA) has been developed~\cite{Mersmann2011}. ELA encompasses several different techniques to quantify the landscape of a black-box problem with numerical surrogates. These range from summary statistics as well as model coefficients to distance-based metrics. The aforementioned calculations are based on a so-called `initial design' $\mathbf{S}$ which essentially is a small sample of the search space $\mathbf{X}$ and the corresponding objective values $\mathbf{Y}$.
The size of this initial design usually depends linearly upon the dimensionality, $D$, of a given problem instance and a common choice is $50D$. An overview of the general procedure of ELA feature generation is depicted in Figure~\ref{fig:ela}.

Two prominent software packages exist to calculate ELA features, namely the R package \texttt{flacco}~\cite{KerschkeT2019flacco} and the Python package \texttt{pflacco}~\cite{pflacco}. These software packages also include the advances made since the inception of ELA, i.e., they also include feature sets proposed over the last decade such as the works of~\cite{funnelkerschke2015} and \cite{munoz2015_ic}.
In this study, we use the same ELA feature sets as used by~\cite{prager2024mvp}, i.e. $y$-distribution, meta model, dispersion, and information content. Note that we adhere to the recommendation of normalizing the objective space prior to feature computation \cite{prager23nully} in order to ensure shift and scale invariance. 

The original work pertaining to ELA features covered several feature sets of which two are used in this study. These are meta model and $y$-distribution with nine and three features respectively. The feature set \textbf{meta model} fits a linear and quadratic model (with and without interactions) to the initial design where the model coefficients and the model quality serve as individual features. This provides insight about the degree of linearity and convexity of a given problem instance. The feature set \textbf{$y$-distribution}, on the other hand, only focuses on the objective value distribution of the initial design. In particular, the kurtosis and skewness are measured~\cite{Mersmann2011}.
The \textbf{dispersion} feature set consists of $16$ features. It segments the objective values of the initial design into different subsets based on varying quantile levels, such as 10\% and 25\%. Within each subset, measurements are taken for both mean and median distances in the decision space. This analysis provides valuable insights into the distribution of local optima, indicating whether they are concentrated in specific regions or dispersed throughout the landscape~\cite{lunacek2006}.
Five features are part of \textbf{information content}. To derive this feature set, a sequence of random walks is performed across the initial design. At each step, comparisons are drawn between the current observation and the subsequent one. These comparisons yield diverse metrics aimed at delineating the landscape's attributes, encompassing aspects of smoothness, ruggedness, and neutrality~\cite{munoz2015_ic}.
Identical to information content, the feature set \textbf{nearest better clustering} includes five features that are metrics and ratios computed between two different sets of distances. One of these sets consists of distances of each observation to its nearest neighbor whereas the other set covers distances between observation and its nearest \textit{better} neighbor~\cite{funnelkerschke2015}.

\subsection{Preprocessing Scheme Based on Target-encoding}\label{sec:te}
In~\cite{prager2024mvp}, the authors developed a preprocessing scheme comprising several steps applied to the initial design $\mathbf{S}$. One of these concerns the representation of categorical variables with real-valued numbers. In particular, two encoding methods, namely \textit{one-hot encoding} and \textit{target-encoding} (TE)~\cite{te2001}, were contrasted with each other. The authors recommend the usage of TE as it performed comparatively better than one-hot encoding and requires less computation time.

TE originates from the \textit{machine learning} (ML) community and is typically used to encode categorical variables for learners incapable of processing these types of variables~\cite{te2001}.
TE measures the influence of a given category of a categorical variable  on the target in question, which
in \cite{prager2024mvp} is modeled as the objective value $Y$ of the initial design $\mathbf{S}$. For a given categorical variable $X_i$ and corresponding specific category $j \in X_i$, TE computes the arithmetic mean of the objective values $Y$ and a subset of $Y_j \subseteq Y$ which only contains objective values where $X_i = j$. A weighted mean is calculated based on these two averages and this new value $j'$ replaces the category $j$.
One can think of $j'$ as a value which measures the average deviation from the arithmetic mean of $Y$ for a category $j$.

\subsection{SHAP-encoding}\label{sec:shap} 

Even though~\cite{prager2024mvp} achieve promising results based on TE the sole purpose of the method was to transform any categorical variable to a numerical value without adding new dimensions. Thus, TE does not capture the variable's effect for individual observations. Rather, it aggregates the mean effect of every feasible category of that variable on the chosen target and replaces the categories with the resulting numerical value.
Every distinct category of a categorical variable is, therefore, assigned the same static real-valued number across all its occurrences in the data set.
A more nuanced approach can be achieved using \textit{SHapley Additive exPlanations} (SHAP)~\cite{NIPS2017_7062}. SHAP is a unified framework for the interpretation of ML model predictions and is based on Shapley values~\cite{shapleyvalues}. 
Shapley values were first introduced in the context of cooperative game theory as a means of quantifying the average contribution of every team member in a coalition game. 
This is done by playing the game in all possible coalitions of team members and averaging over the difference in the outcome before a team member joins and after the member has joined a coalition.
This accounts for all interaction effects between the different players and leads to desirable properties like local accuracy, missingness and consistency~\cite{NIPS2017_7062}. 
The SHAP framework transfers this approach to the ML domain by conceptualizing every observation in a dataset as a game. For instance, in the case of tabular data, this means that every row in the dataset is a distinct game while every feature value in that row can be viewed as a player. The game outcome corresponds to the prediction of the ML model of interest. 
The Shapley values can than be calculated according to \refeq{eq:shapley}~\cite{NIPS2017_7062} by
\begin{equation}
    \label{eq:shapley}
    \phi_i(f,x) = \sum_{z'\subseteq x'} \frac{|z'|!(M-|z'|-1)!}{M!}\left[f_x(z')-f_x(z'\setminus i)\right].
\end{equation}
Here, $\phi_i$ is the Shapley value for the feature value expressed by feature $i$ within the observation $x$ (e.g. tabular data row) while being subject to the ML model $f$. $M$ denotes the number of features in the dataset, $x'$ is the \textit{simplified input} which is commonly used by explanation models and can be mapped to the original input by any $h_x(x')=x$. Finally, $z'\in \{0,1\}^M$ denotes the number of features that are part of the respective coalition. In practice, ML models usually exhibit static architectures. This means the number of input features cannot be arbitrarily varied. To overcome this hurdle every feature that is not part of a coalition instance is sampled randomly from all values exhibited by it within the entire dataset. The idea is to randomize the feature and, thus, make it lose its predictive power. The SHAP framework also provides several solutions for the NP-hardness of the Shapley value calculation exhibited by \refeq{eq:shapley}. The current default relies on antithetic sampling which considers an entire permutation of all features in forward and reverse direction~\cite{antithetic2022}.
Thus, SHAP allows to calculate Shapley values even for a large number of features. This provides a local (for a single observation) and additive quantization of the contribution of every feature to the deviation from the expected value. Here, the expected value is the prediction if no feature is part of the coalition. 
As SHAP values are synonymous to Shapley values applied in the ML domain, we refer to them as SHAP values.
In this work, we train a ML model on the initial design $\mathbf{S}$ of ELA where the input of the model is the sample of decision space $X$ and corresponding objective values $Y$. For each individual observation in $X$, we now compute the SHAP values of all \textbf{categorical} decision variables on the prediction (i.e., the objective value) of the model. We then replace every category in this specific observation with the respective SHAP values. This process is repeated until every observation in $X$ is iterated over and all non-numeric values are replaced.

\section{Experimental Setup}\label{sec:exp-setup}

\subsection{Benchmark Problems}\label{sec:yahpo}
\begin{table}[htbp]
    \centering
    \caption{Overview of available scenarios in YAHPO Gym, the search space, the number of problem instances per scenario and information whether the search domain is hierarchical or not. 
    }
    \label{tab:yahpo}
    \begin{tabular}{llll}
        \toprule
        Scenario & Search Space (cont, int, cat) & \# Instances & H \\
        \midrule
        rbv2\_glmnet & 3D: (2, 0, 1) & 115 & --\\
        rbv2\_rpart & 5D: (1, 3, 1) & 117 & --\\
        rbv2\_aknn & 6D: (0, 4, 2) & 118 & --\\
        rbv2\_svm & 6D: (3, 1, 2) & 106 & $\checkmark$ \\
        iaml\_ranger & 8D: (2, 3, 3) & 4 & $\checkmark$ \\
        rbv2\_ranger & 8D: (2, 3, 3) & 119 & $\checkmark$ \\
        iaml\_xgboost & 13D: (10, 2, 1) & 4 & $\checkmark$ \\
        rbv2\_xgboost & 14D: (10, 2, 2) & 119 & $\checkmark$ \\
        \bottomrule
    \end{tabular}
\end{table}
We use HPO problems as representatives of the in general broader class of MVP problems which also e.g. prominently comprises engineering applications such as aircraft design~\cite{10.1007/978-3-319-67988-4_140,Venter2004} and design of induction motors~\cite{Liuzzi2005}.   The HPO problem set is provided by the  'Yet Another Hyperparameter Optimization Gym' (YAHPO Gym)~\cite{yahpo} benchmark.
All considered HPO problems are classification tasks and  associated with a specific learner, its respective hyperparameters that need to be tuned and final metric which represents quality of the learner on this given training task. In the context of YAHPO Gym, a scenario comprises several HPO problems for a specific learner. The different problems within a scenario are instantiated using various datasets sourced from OpenML~\cite{OpenML2013}, with the objective metric being the misclassification error, which we aim to minimize.

For the sake of comparison, we use the same subset of scenarios as the authors of~\cite{prager2024mvp}. These are presented in~\reftab{tab:yahpo}, which delineates the dimensions of the search space and the distribution of continuous, integer, and categorical decision variables. In total, this encompasses $702$ individual problem instances. 
It is essential to emphasize that the instances within a scenario may not necessarily share similar levels of difficulty or overall landscape characteristics. Factors such as the presence of multiple peaks or the unimodality of the fitness landscape are contingent not only upon the scenario but also the specific dataset employed. Thus, the categorization outlined in~\reftab{tab:yahpo} serves solely for illustrative purposes and should not be considered as indicative of problem similarity. 

\subsection{Algorithm Portfolio}\label{sec:portfolio}

The performance data of our algorithm portfolio was provided by the authors of~\cite{prager2024mvp} which benefits comparability. The circumstances under which this data was created are briefly described in the following. \texttt{SMAC3}~\cite{Lindauer_SMAC3_A_Versatile_2022}, \texttt{Optuna}~\cite{optuna_2019}, \texttt{pymoo}~\cite{pymoo}, and random search (RS) were compared on the previously described benchmark functions.
The allotted budget of each algorithm amounts to $100D$ function evaluations where $D$ denotes the problem dimensionality.
Each algorithm is executed on each problem instance of YAHPO Gym $20$ times. The runtime of these $20$ repetitions of an algorithm until a predefined target is reached is aggregated into a single value called \textit{expected running time} (ERT)~\cite{ert}.




A target can be perceived as a threshold from which ontowards we deem the achieved results satisfactory. In this particular setting, YAHPO Gym models HPO problems with an underlying classification task. The chosen metric to optimize for is the misclassification error.
In theory, the best performance is achieved with a misclassification error of zero. However, in practice this might not be reachable as this is constrained by the data set, the chosen ML model and the box-constraints for its hyperparameters. Hence, \cite{prager2024mvp} determine an individual target for each problem instance. This target is calculated by (1) concatenating all optimizer traces of the four algorithms (including their $20$ repetitions), (2) ordering the objective values in ascending order, and (3) selecting the $0.01$-quantile of the objective values. Thereby, the target is chosen in a manner that it is still challenging for all four algorithms to achieve but simultaneously guarantees that at least one algorithm is able to solve each individual problem instance.


\subsection{Exploratory Landscape Feature Generation}\label{sec:exp-ela}
While we use the data provided by \cite{prager2024mvp} for the target-encoded ELA features, we generate the ELA feature values based on our proposed SHAP-encoding by using the Python package \texttt{pflacco}~\cite{pflacco}.
The initial design $\mathbf{S}$ is generated uniformly at random in the feasible domain of each problem, with a sample size of $50D$. Again, $D$ represents the dimensionality of any given problem instance. Since the calculation of SHAP values for a single prediction has a complexity of $O(2^N)$, we make use of the \texttt{PermutationExplainer} from the Python package provided with the \texttt{SHAP} publication~\cite{NIPS2017_7062}. This is the current default \texttt{Explainer} of the package and employs antithetic sampling to calculate the SHAP values~\cite{antithetic2022}. The ML model, which serves as basis to calculate the respective SHAP values, is a random forest with a default configuration of the Python package \texttt{scikit-learn}~\cite{scikit-learn}. 

After replacing each non-numeric value in the initial design $\mathbf{S}$ with its corresponding SHAP value, we normalize each decision variable and the objective value of a given initial design independently, constraining them within the interval $[0, 1]$ as recommended in ~\cite{prager23nully,prager2024mvp}. Each set of ELA features is calculated $20$ times for a given problem instance to produce more robust results.\footnote{Corresponding source code and results can be found on \url{https://github.com/konsdt/PPSN-SHAP-TE-ELA}} 

\subsection{Construction of Algorithm Selectors}\label{sec:exp-aas}

Our approach conceptualizes the AAS \cite{KerschkeHNT2019AutomatedAlgorithm} scenario as a multi-class classification problem. The ELA features serve as input for the AS model while the class label is determined by the best-performing algorithm (lowest ERT value) for each particular instance.
We construct two AAS models, one trained on target-encoded and one trained on SHAP-encoded ELA features.
It is pertinent to mention that our dataset encompasses a total of $14,040$ observations for each encoding variant, resulting from the generation of ELA features from 20 different samples per problem instance (of which there are 702). The class label per instance remains consistent across these 20 repetitions.

For the sake of comparability, we use the same ML model as used by~\cite{prager2024mvp}, i.e., a random forest provided by the Python package \texttt{scikit-learn}~\cite{scikit-learn}.

We assess the model through a 10-fold cross-validation approach, ensuring that all repetitions of a specific problem instance are grouped within a single fold. This strategy prevents the dispersion of repetitions across multiple folds, which could otherwise result in instances being both trained and tested simultaneously. Such an arrangement helps maintain the integrity of the evaluation process, minimizing potential biases and ensuring robustness in our analysis.

As delineated below, the two models exhibit complementary behavior, prompting us to subsequently propose two strategies that capitalize on this synergy.

\section{Results}\label{sec:results}
\begin{figure}[ht!]
    \centering
    \includegraphics[width=\textwidth]{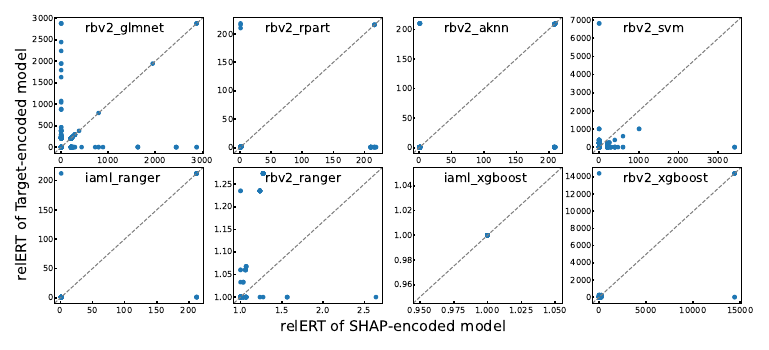}
    \caption{Performance of both algorithm selectors. The $x$-axis shows the relERT values of the SH model whereas the $y$-axis shows the relERT values of the TE model. Points on the grey line exhibit (nearly) identical relERT values for both models. Points above the grey line represent instances where the SH model produces a better performance. Points below the grey line represent the other case.}
    \label{fig:performance-scatter}
\end{figure}

The comparison of our SHAP value based algorithm selector (SH model) contrasted against the target-encoded based algorithm selector (TE model) shows a very similar performance between both models as reflected by the TE and SH columns in~\reftab{tab:performance_VBS}.
A closer look into the performance distribution compartmentalized into the respective YAHPO Gym scenarios provides a more insightful view. This is shown in~\reffig{fig:performance-scatter}. The figure depicts the performances of both models for any given problem instance where the $x$-coordinate is determined by the relERT value of the SH model and the $y$-coordinate by the relERT value of the TE model respectively. Points on the gray line represent instances where the performance of both models is either identical or very similar. The relERT values are determined by dividing all ERT values of a single problem instance by the ERT value of the virtual best solver (VBS).
The VBS embodies a theoretical algorithm selector that unfailingly selects the optimal algorithm from our portfolio for every instance.

Again, we can discern that the distribution and consequently summary statistics like the mean performance of the models are similar. Yet, we can also observe that there are a multitude of problem instances where the SH model excels while the TE model exhibits a poor performance and vice versa. For example, this dichotomy is present in the scenario \texttt{rbv2\_svm}, where points in the bottom left show this behavior. At the same time there are also points in the top left and bottom right corner. Especially, the latter contribute significantly to the complementarity of both models since in these cases one of the models manages to select the VBS while the other selects the worst solver from the portfolio.

\begin{figure}[ht!]
    \centering
    \includegraphics[width=1\textwidth]{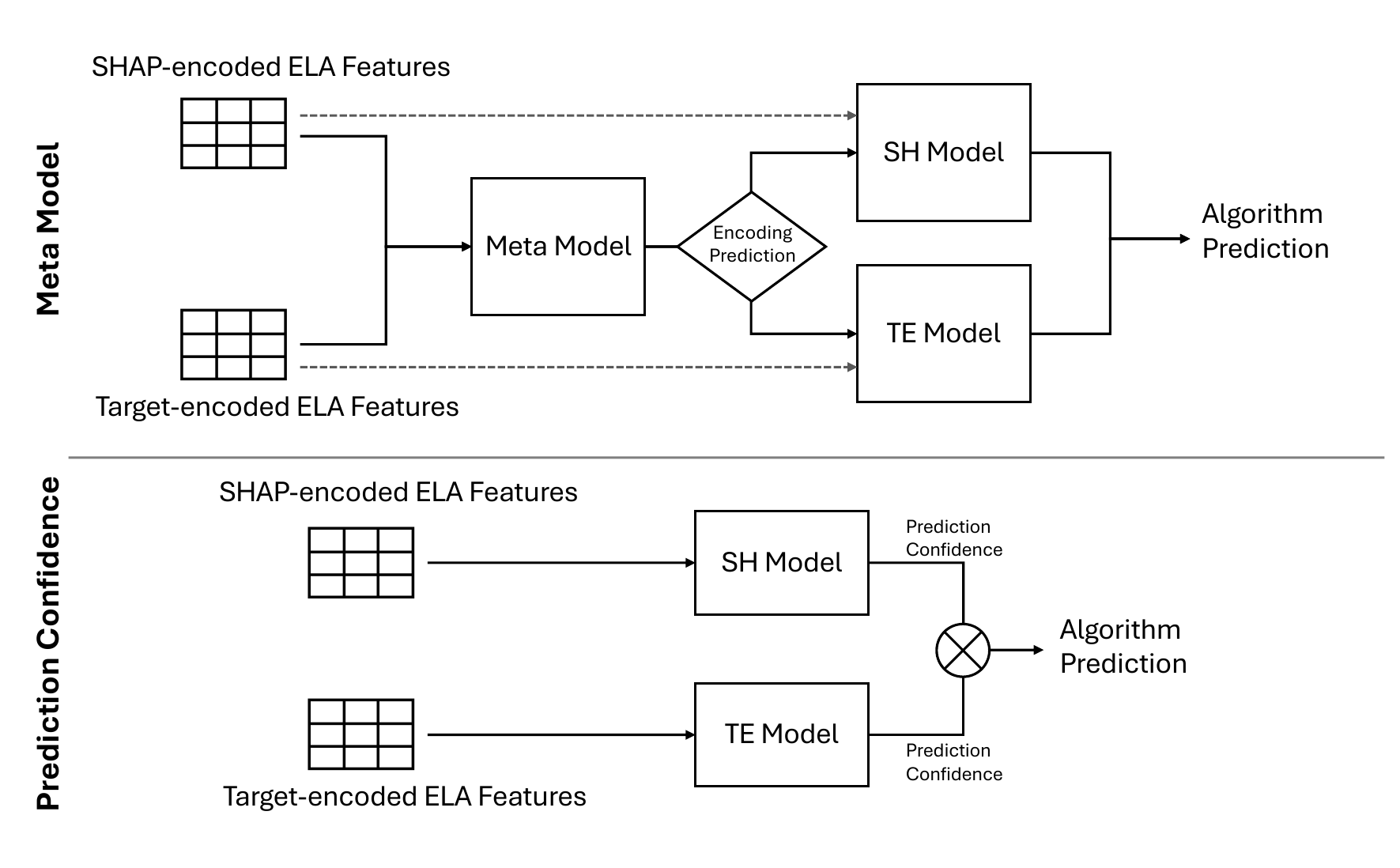}
    \caption{Illustration of our two suggested approaches to capitalize on the complementary of the two AAS strategies. The meta model classifier utilizes ELA features derived from SHAP values and TE for each problem instance. With a binary target variable representing either the TE or SH model based on better prediction performance, the classifier selects the appropriate algorithm selector. Subsequently, the chosen selector utilizes ELA features specific to its encoding type and selects an algorithm from the portfolio. Alternatively, the prediction confidence approach relies on comparing prediction probabilities from independent AAS models, with the higher probability indicating greater confidence in the prediction.}
    \label{fig:meta-model}
\end{figure}
Even after careful and rigorous analysis we cannot identify the reasons that lead to this dichotomous state. Nevertheless, exploiting this complementary behavior of both encoding methods produces an overall better result.
The full potential of hybridizing both encoding variants can be seen in the column \textbf{'Hybrid'} in table~\reftab{tab:performance_VBS}. The values shown there are determined by always selecting the superior algorithm between the two predicted by the TE and the SH model. 
Looking at the relERT averaged over all problem instances we see that even the hybrid approach still performs $10.82$ times worse than the virtual best solver (VBS) of the selected portfolio.
To bring this into context we also report the relERT values of the single best solver (SBS) which, as already reported in~\cite{prager2024mvp}, is \texttt{SMAC3}. In agreement with~\cite{prager2024mvp} we see that algorithm selection exhibits a large improvement over the general selection of a single solver for all problems.
Even though there remains room for improvement the hybridization is a substantial advancement over the individual approaches as it bears $40.42\%$ performance increase over the SH model and $39.49\%$ over the TE model. 
To leverage this relationship we suggest two approaches which are conceptualized in~\reffig{fig:meta-model}. In the first approach that can be seen in the top of~\reffig{fig:meta-model}, we utilize an additional ML model which is responsible for determining which of the two algorithm selectors to use. We call this new model \textbf{`meta model'}.

The `meta model' approach is based on a random forest classifier that receives both, the ELA features based on SHAP values and the ELA features based on TE for every problem instance respectively. The target variable has two classes, and can either be the TE model or the SH model, depending on which was able to predict the better solver for the problem.
It then decides which of the two algorithm selectors it should use. Thereafter, the chosen algorithm selector uses only the ELA features according to its respective encoding type and chooses an appropriate algorithm out of the portfolio.

The second approach relies solely on the comparison of prediction probabilities between the two independent AS models. We evaluate both selectors at the same time and trust the prediction which exhibits a higher prediction probability. Doing so, we interpret the prediction probability as prediction confidence of the respective model and thus denote this approach as \textbf{`Confidence'}. For a multiclass random forest classifier the predicted probabilities for each class of an input sample are determined by averaging the predicted probabilities across all trees within the forest. In each individual tree, the class probability is calculated as the proportion of samples belonging to the same class within a leaf node. A high-level representation of that concept is illustrated in the bottom of~\reffig{fig:meta-model}.
We asses both approaches using 10-fold cross validation and the results are shown in the columns labeled `Meta' and `Confidence' of~\reftab{tab:performance_VBS}. The highlighted values mark the best selection strategy for the respective scenario. We find that the prediction confidence approach performs the best across all problem instances. When examining individual scenarios, the selection method relying on prediction confidence is outperformed by the alternative meta-model approach in only two cases. In the two instances where its performance aligns with the meta-model approach, there is little to no potential for enhancement as the SBS equals or nearly equals the VBS.
\begin{table}[ht!]
\centering
\caption{Performance of the different approaches which is measured as the arithmetic mean of the relERT value. Highlighted values indicate the approaches that performed best for a given setting. The column `Hybrid' is not considered since it represents a virtual best encoding choice. }
\label{tab:performance_VBS}
\begin{tabular}{ll|c|cc|c|cc}
  \toprule
\textbf{Scenario} &  \begin{tabular}{c}
   \textbf{\# Instances $\times$}    \\
   \textbf{\# samples} 
\end{tabular} & \textbf{SBS} & \textbf{TE} & \textbf{SH} & \textbf{Hybrid} & \textbf{Meta} & \textbf{Confidence} \\ 
\midrule rbv2\_glmnet  & 2\,120 & 558.45 & 50.90 & 47.05  & 29.08  & 45.35  & \cellcolor[HTML]{CDCDE6} 44.90 \\ 
\midrule rbv2\_rpart   &   2380 & 28.96  & \cellcolor[HTML]{CDCDE6} 1.66  & 3.11  & 1.39   & 1.76   & 2.20\\ 
\midrule rbv2\_aknn    & 2\,360 & 4.54   & 2.77  & 3.29  & 2.15  & \cellcolor[HTML]{CDCDE6}2.32  & 2.85\\
\midrule rbv2\_svm     &     80 & 330.86 & 19.18 & 15.80  & 4.89  & 14.39  & \cellcolor[HTML]{CDCDE6}10.47 \\ 
\midrule iaml\_ranger  & 2\,300 & 53.85  & \cellcolor[HTML]{CDCDE6}40.63 & 51.20 & 37.99 & \cellcolor[HTML]{CDCDE6}40.63 & 45.92\\ 
\midrule rbv2\_ranger  &     80 & 1.02   & \cellcolor[HTML]{CDCDE6}1.01  &\cellcolor[HTML]{CDCDE6} 1.01   & 1.01   & \cellcolor[HTML]{CDCDE6}1.01   & \cellcolor[HTML]{CDCDE6}1.01 \\ 
\midrule iaml\_xgboost & 2\,340 & \cellcolor[HTML]{CDCDE6}1.00   &\cellcolor[HTML]{CDCDE6} 1.00  & \cellcolor[HTML]{CDCDE6}1.00  & 1.00  & \cellcolor[HTML]{CDCDE6}1.00  & \cellcolor[HTML]{CDCDE6}1.00\\
\midrule rbv2\_xgboost & 2\,380 & 127.87 & 32.44 & 38.48  & 25.57  & 32.36  &\cellcolor[HTML]{CDCDE6} 26.30 \\ 
   
\midrule \midrule All & 14\,040 & 169.19 & 17.88 & 18.16  & 10.82  & 16.18 & \cellcolor[HTML]{CDCDE6} 14.68 \\ 
\midrule  \bottomrule
\end{tabular}
\end{table}
\begin{table}[ht!]
\centering
\caption{Performance of the different approaches which is measured as the arithmetic mean of the relERT normalized with respect to the VBE given by the `Hybrid' column.} 
\label{tab:performance_VBE}
\begin{tabular}{ll|cc|c|cc|c}
  \toprule
\textbf{Scenario} & \begin{tabular}{c}
   \textbf{\# Instances $\times$}    \\
   \textbf{\# samples} 
\end{tabular} & \textbf{TE} & \textbf{SH} & \textbf{Hybrid} & \textbf{Meta} & \textbf{Confidence} \\ 
\midrule rbv2\_glmnet  & 2\,120 & 22.50  & 17.95  & 1.00  &\cellcolor[HTML]{CDCDE6} 16.31 & 16.89  \\ 
\midrule rbv2\_rpart   &   2380 &\cellcolor[HTML]{CDCDE6} 1.28   & 2.72  & 1.00   & 1.37   & 1.82  \\ 
\midrule rbv2\_aknn    & 2\,360 & 1.62  & 2.15  & 1.00  &\cellcolor[HTML]{CDCDE6} 1.18  & 1.71  \\
\midrule rbv2\_svm     &     80 & 15.28  & 11.90  & 1.00  & 10.50  &\cellcolor[HTML]{CDCDE6} 7.36  \\ 
\midrule iaml\_ranger  & 2\,300 &\cellcolor[HTML]{CDCDE6} 3.64 & 14.21 & 1.00	 &\cellcolor[HTML]{CDCDE6} 3.64 & 8.93  \\ 
\midrule rbv2\_ranger  &     80 &\cellcolor[HTML]{CDCDE6} 1.00   &\cellcolor[HTML]{CDCDE6} 1.00   & 1.00   &\cellcolor[HTML]{CDCDE6} 1.00   &\cellcolor[HTML]{CDCDE6} 1.00  \\ 
\midrule iaml\_xgboost & 2\,340 &\cellcolor[HTML]{CDCDE6} 1.00  &\cellcolor[HTML]{CDCDE6} 1.00  & 1.00  &\cellcolor[HTML]{CDCDE6} 1.00  &\cellcolor[HTML]{CDCDE6} 1.00 \\
\midrule rbv2\_xgboost & 2\,380 & 7.88  & 13.92  & 1.00  & 7.78  &\cellcolor[HTML]{CDCDE6} 1.73  \\ 
   
\midrule \midrule All & 14\,040 & 8.01  & 8.17  & 1.00  & 6.20 &\cellcolor[HTML]{CDCDE6} 4.99\\ 
\midrule  \bottomrule
\end{tabular}
\end{table}

To further illustrate how well both approaches perform we lean on the idea of relERT calculation. Since the values in the `Hybrid' column of~\reftab{tab:performance_VBS} result from always choosing the best feature encoding strategy we consider it to be the \textbf{`virtual best encoding' (VBE)}. Hence, we normalize all relERT values by dividing with the VBE values. The resulting values are shown in~\reftab{tab:performance_VBE}. Continuing this analogy, the \textbf{`single best encoding' (SBE)} from the two compared encoding methods is target-encoding. 
We refrain from showing the SBS column in~\reftab{tab:performance_VBE} since it reveals no additional relevant information. 
In total, the prediction confidence approach is able to close the gap between SBE and VBE by $43.10\%$. The meta model also achieves an overall improvement over the SBE but this only amounts to a gap closure of $25.84\%$.

\section{Conclusion}\label{sec:conclusion}

Landscape analysis has been used in a variety of studies with different objectives and scopes over the last decade. However, it has been largely limited to either the continuous or combinatorial optimization domain. Recent advances of~\cite{wmodel}, \cite{prager23mip}, \cite{prager2024mvp} have iteratively extended this to the domain of mixed-variable problems.
One of the main contributions of~\cite{prager2024mvp} is the proposition to transform categorical variables into continuous representations. This is achieved by employing techniques such as one-hot encoding and target-encoding.

In our work, we extend this particular component by introducing SHAP values as an alternative encoding method to transform categorical into continuous values. We demonstrate the merits of our encoding variant in an automated algorithm selection setting and compare our results with the results of~\cite{prager2024mvp}.
While we cannot determine any substantial difference between the algorithm selector based on the SHAP-encoding and the one based on target-encoding, we show that both encoding methods are performance complementary and achieve a much better performance when switching between them. We propose two strategies to hybridize both encoding methods. 
One devised approach uses a meta model that decides which subsequent algorithm selector to use. The other compares the prediction probabilities of the AS models trained on both encodings and interprets them as prediction confidence.
Both approaches are able to substantially improve the existing results based on target-encoding. In our setting these results are considered the single best encoding while the hybridization of both discussed encodings is considered the virtual best encoding. While the meta model closes the gap between both by $25.84\%$, the prediction confidence method is able to achieve a substantial improvement with a gap closure of $43.10\%$.

Despite this advancement, there remains room for improvement which becomes especially clear when considering that the relERT value of the virtual best encoding is still quite high with $10.82$.
At the forefront of research questions unsolved is, what the underlying mechanisms are which lead to the performance complementary behavior of these two encoding variants. Revealing these will foster our understanding and will help us to incorporate a more sophisticated encoding mechanism directly into the ELA feature calculation.
We also plan to further investigate if different ML models during SHAP value calculation can change the expressiveness of the SHAP-encoded features and will also extend our scope to MVP problems beyond the so far considered subclass of HPO problems.

\subsubsection{\ackname} This work was realized with the financial support of ANR project ANR-22-ERCS-0003-01 and of CNRS Sciences informatiques project \emph{IOHprofiler}.

%
%
%
\bibliographystyle{splncs04}
\bibliography{99_references}

\end{document}